\UseRawInputEncoding
\documentclass{article}

\usepackage{microtype}
\usepackage{graphicx}
\usepackage{subcaption}
\usepackage{booktabs} 

\usepackage{hyperref}

\usepackage[preprint]{icml2026}

\usepackage{amsmath}
\usepackage{amssymb}
\usepackage{mathtools}
\usepackage{amsthm}
\usepackage{dsfont} 

\usepackage[capitalize,noabbrev]{cleveref}

\usepackage[most]{tcolorbox}
\tcbset{promptboxstyle/.style={
  enhanced,
  breakable,
  colback=violet!4,
  colframe=violet!30,
  boxrule=0.6pt,
  arc=2pt,
  left=6pt,
  right=6pt,
  top=4pt,
  bottom=4pt,
  fonttitle=\bfseries,
  coltitle=black
}}
\newtcblisting{promptbox}[2][]{
  promptboxstyle,
  title={#2},
  listing only,
  listing options={
    basicstyle=\ttfamily\footnotesize,
    breaklines=true,
    columns=fullflexible,
    keepspaces=true,
    showstringspaces=false
  },
  #1
}

\theoremstyle{plain}

\theoremstyle{definition}

\theoremstyle{remark}

\usepackage[textsize=tiny]{todonotes}

\icmltitlerunning{DoubleTake: Contrastive Reasoning for Faithful Decision-Making in Medical Imaging}

\begin{document}

\twocolumn[
  \icmltitle{DoubleTake: Contrastive Reasoning for Faithful Decision-Making\\
     in Medical Imaging }



    \icmlsetsymbol{equal}{*}
    
    \begin{icmlauthorlist}
      \icmlauthor{Daivik Patel}{equal,ru}
      \icmlauthor{Shrenik Patel}{equal,ru}
    \end{icmlauthorlist}
    
    \icmlaffiliation{ru}{Rutgers University}
    
    \icmlcorrespondingauthor{Daivik Patel}{daivik.d.patel@rutgers.edu}
    \icmlcorrespondingauthor{Shrenik Patel}{shrenik.patel@rutgers.edu}
  \icmlkeywords{Machine Learning, ICML}

  \vskip 0.3in
]


\printAffiliationsAndNotice{\icmlEqualContribution}  

\begin{abstract}
Accurate decision-making in medical imaging requires reasoning over subtle visual differences between confusable conditions, yet most existing approaches rely on nearest-neighbor retrieval that returns redundant evidence and reinforces a single hypothesis. We introduce a \textbf{contrastive, document-aware reference selection framework} that constructs compact evidence sets optimized for discrimination rather than similarity by explicitly balancing visual relevance, embedding diversity, and source-level provenance using ROCO embeddings and metadata. While ROCO provides large-scale image--caption pairs, it does not specify how references should be selected for contrastive reasoning, and naive retrieval frequently yields near-duplicate figures from the same document. To address this gap, we release a \textbf{reproducible reference selection protocol and curated reference bank} that enable a systematic study of contrastive retrieval in medical image reasoning. Building on these contrastive evidence sets, we propose \textbf{Counterfactual-Contrastive Inference (CCI)}, a confidence-aware reasoning framework that performs structured pairwise visual comparisons and aggregates evidence using margin-based decision rules with faithful abstention. On the MediConfusion benchmark, our approach achieves \textbf{state-of-the-art performance}, improving \textbf{set-level accuracy by more than 15\%} relative to prior methods while substantially reducing confusion and maintaining high conditional accuracy.
\end{abstract}

\section{Introduction}

Medical image decision-making is often framed as a recognition problem, yet many of the most consequential errors arise after recognition has already succeeded \cite{topol2019high}. In clinical practice, failures frequently occur not because a finding is missed, but because two plausible diagnoses are confused. These errors emerge in regimes where images share global structure, modality, and anatomical context, and differ only in subtle, localized visual cues. In such settings, the core challenge is not identifying what is present, but determining which of several visually similar hypotheses is better supported by the evidence.

Benchmarks such as MediConfusion are designed to isolate this challenge \cite{sepehri2024mediconfusion}. Rather than evaluating images independently, MediConfusion organizes data into pairs of visually confusable images that require different answers. This structure exposes a class of failures that conventional metrics often obscure. A model may achieve strong image-level accuracy while repeatedly assigning the same prediction to both images in a pair. From a clinical perspective, such behavior is unacceptable: confusing look-alike conditions is precisely the error that decision support systems are meant to avoid. Set-level accuracy, which requires both images in a confusion pair to be classified correctly, therefore provides a more stringent and clinically aligned measure of discriminative competence.

Recent advances in vision-language modeling suggest that performance in medical imaging is increasingly limited by reasoning under ambiguity rather than coarse recognition. In confusion-heavy regimes, systems often collapse toward a single answer due to shortcut correlations or dominant priors, yielding high confusion rates despite reasonable marginal accuracy.

Retrieval augmentation has emerged as a common strategy for improving medical image reasoning, motivated by the intuition that relevant examples can provide useful context \cite{lewis2020rag}. In practice, retrieval is almost universally implemented through nearest-neighbor search in an embedding space, selecting the most similar images to the query. While effective for recall-oriented tasks, this approach is poorly suited to discriminative decision making, where evidence diversity is as important as relevance \cite{carbonell1998mmr}. Nearest-neighbor retrieval optimizes for similarity rather than contrast, and frequently returns near-duplicate images that share diagnosis, visual structure, and even document provenance. In medical corpora derived from the literature, this redundancy is exacerbated by multiple figures from the same article depicting the same condition. Rather than clarifying decision boundaries, such evidence reinforces a single hypothesis and suppresses alternatives, increasing decisiveness without increasing correctness. Addressing this misalignment requires rethinking retrieval not as a similarity operation, but as a process of evidence construction tailored to discrimination.

The main contributions of this work are as follows.

\noindent\textbf{(1)} We propose a \textbf{contrastive, document-aware reference selection framework} that constructs compact evidence sets optimized for \emph{discrimination rather than similarity} by explicitly balancing visual relevance, embedding diversity, and source-level provenance. To support reproducibility and systematic evaluation, we release a \textbf{triad selection protocol and curated reference bank} derived from ROCO, enabling controlled study of contrastive retrieval in medical image reasoning.

\noindent\textbf{(2)} Building on these contrastive evidence sets, we introduce \textbf{Counterfactual-Contrastive Inference (CCI)}, a \textbf{confidence-aware reasoning framework} that aggregates structured pairwise visual comparisons using margin-based decision rules with faithful abstention. We further introduce a pair-level adjudication mechanism to resolve ambiguity in confusion-pair settings without imposing artificial label constraints. Together, these components achieve \textbf{state-of-the-art performance on MediConfusion}, substantially improving set-level accuracy while reducing confusion rates.

\begin{figure*}[t]
  \centering
  \includegraphics[width=0.99\textwidth]{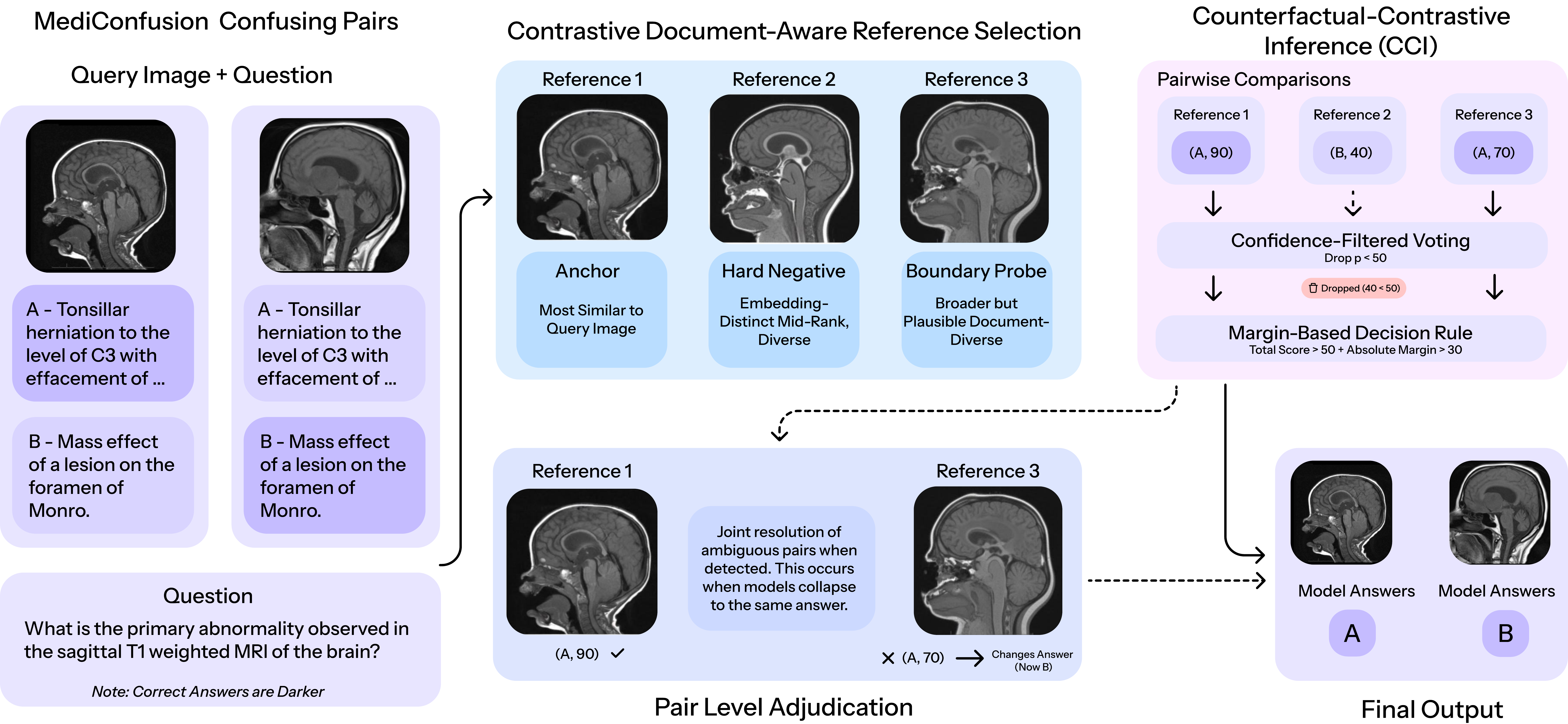}
  \caption{
    Overview of the proposed framework. Given a query image, a contrastive triad of reference images
    is retrieved from an external medical image bank. Structured pairwise comparisons are aggregated
    by Counterfactual-Contrastive Inference (CCI) using confidence-aware voting, margin-based decisions,
    and adjudication for ambiguous cases.
  }
  \label{fig:system_overview}
\end{figure*}

\section{Related Work}

\subsection{Medical Visual Question Answering Benchmarks}
Medical visual question answering (VQA) has been studied through a range of benchmarks, including ImageCLEF VQA-Med \cite{hasan2018imageclef_vqamed,abacha2019vqa}, PathVQA \cite{he2020pathvqa}, and knowledge-enhanced datasets such as SLAKE \cite{liu2021slake}.

\subsection{Retrieval for Medical Image Reasoning}
Retrieval augmentation is commonly used in medical image reasoning to provide contextual exemplars when decisions hinge on subtle visual cues. Most existing systems implement retrieval via similarity search in an embedding space, returning the top-$k$ nearest neighbors of a query image, sometimes augmented with captions or metadata. While effective for recall-oriented tasks, this similarity-first objective is poorly aligned with confusion-style benchmarks such as MediConfusion, where the goal is to \emph{discriminate} between visually similar conditions rather than retrieve more of the same.

This misalignment is amplified in literature-derived corpora, where figures are often templated and multiple near-duplicate panels originate from the same document. As a result, naive top-$k$ retrieval frequently returns redundant, provenance-collapsed evidence that reinforces a single hypothesis instead of exposing alternatives. Our work addresses this gap by framing retrieval as \emph{evidence construction for discrimination}, with explicit controls for diversity and document provenance.

\subsection{Contrastive Reasoning at Inference Time}
The term \emph{contrastive} is typically used to describe representation learning objectives for medical vision--language alignment, for example via CLIP-style pretraining and medical variants such as MedCLIP \cite{wang2022medclip}. More recent unified medical CLIP-style models include UniMed-CLIP \cite{khattak2024unimedclip}. These approaches improve representation quality, including in zero-shot settings \cite{tiu2022chexzero,jang2022zero}, but they do not by themselves specify how retrieved evidence should be selected for discriminative decision making.

Our framing is orthogonal: we do not introduce a new contrastive training objective. Instead, we construct \emph{contrastive evidence sets at inference time}, forcing the model to compare the query against carefully selected alternatives that probe decision boundaries. This perspective complements recent instruction-tuned and few-shot medical multimodal systems \cite{zhang2023pmc-vqa,moor2023medflamingo} as well as earlier vision--language baselines \cite{li2019visualbert,vaswani2017attention}.

\subsection{Reliability via Abstention + Evidence Aggregation}
Selective prediction and abstention are well-established tools for improving reliability in high-stakes settings, allowing models to defer decisions when evidence is insufficient \cite{geifman2017selective,angelopoulos2021gentle}. In medical imaging, this is particularly important because multiple visually plausible hypotheses often coexist.

Confusion-pair benchmarks such as MediConfusion make this challenge explicit by evaluating whether models can distinguish look-alike conditions rather than merely achieve high marginal accuracy. In this setting, decisions arise from multiple structured visual comparisons whose confidence and consistency may vary and can be miscalibrated \cite{guo2017calibration}, and naive aggregation can collapse to identical predictions across a pair. We address these failure modes with confidence-aware aggregation and faithful abstention.

\section{Method}
\label{sec:method}

\subsection{System Overview}
We present a two-stage inference framework for discriminative medical image decision making. Given a query image $x$ and a binary multiple-choice question with two candidate answer options $\mathcal{Y}=\{A,B\}$ (where $A$ and $B$ denote the two answer \emph{strings} provided by the benchmark for that question), the system outputs a prediction $\hat{y}\in\{A,B,\bot\}$, where $\bot$ denotes abstention. The framework combines \textbf{contrastive, document-aware reference selection}, which constructs compact evidence sets optimized for discrimination, with \textbf{Counterfactual-Contrastive Inference (CCI)}, which aggregates structured visual comparisons into a reliable decision.

For a given query image $x$, the reference selection component retrieves a fixed triad of reference images $R(x)=\{r_1,r_2,r_3\}$ from a large external medical image bank. Each reference induces a counterfactual comparison that probes how the decision would change relative to a plausible alternative, producing a grounded vote with an associated confidence. CCI filters unreliable signals, aggregates evidence using a margin-based rule, and invokes adjudication only when evidence is ambiguous. In datasets organized into visually confusable pairs, a final pair-level adjudicator resolves ambiguous outcomes. Figure~\ref{fig:system_overview} provides an overview of the full pipeline.

\subsection{Contrastive, Document-Aware Reference Selection}

Effective discrimination requires evidence that is not only relevant to the query, but also diverse and non-redundant. Standard nearest-neighbor retrieval optimizes for similarity and frequently returns near-duplicate images, often originating from the same document, which reinforces a single hypothesis and obscures subtle distinctions. We therefore design reference selection explicitly as \emph{evidence construction for discrimination}.

\textbf{Reference bank and representations.}
We construct a reference bank derived from ROCO that provides a large, redundant universe of medical images with document provenance \cite{pelka2018roco}. Formally, let
\[
\mathcal{B}=\{(r_j,c_j,d_j,m_j)\}_{j=1}^N
\]
denote the reference bank, where $r_j$ is an image, $c_j$ its caption, $d_j$ a document identifier (e.g., PMC article ID), and $m_j$ its imaging modality. All images are embedded using a frozen encoder $\phi(\cdot)$ (CLIP ViT-B/32) \cite{radford2021learning,dosovitskiy2020vit}. For a query image $x$, cosine similarity is defined as
\[
s(x,r)=\frac{\langle \phi(x),\phi(r)\rangle}{\|\phi(x)\|\|\phi(r)\|}.
\]
Because ROCO contains many visually redundant figures, we apply near-duplicate suppression. A candidate $r$ is removed if there exists a previously retained image $r'$ such that
\[
s(r,r')>\tau_{\mathrm{dup}}, \qquad \tau_{\mathrm{dup}}=0.99.
\]
This prevents trivial duplicates from dominating the retrieved set.

\textbf{Candidate pool and rank bands.}
Let $\mathcal{C}(x)$ denote the candidate pool after modality gating and deduplication (defined below). We rank all $r\in\mathcal{C}(x)$ in descending order of similarity $s(x,r)$ and write $\mathcal{C}_{[a,b]}(x)$ for the subset of candidates whose similarity rank lies in $[a,b]$ (1-indexed, inclusive). When selecting $r_2$ and $r_3$, we additionally exclude any already selected references and, whenever possible, enforce distinct document IDs.

\textbf{Triad retrieval policy.}
We retrieve exactly three references per query. This fixed budget encourages diversity and stabilizes downstream inference. The three references play complementary roles.

The \emph{anchor reference} provides a stable visual baseline and is selected as the most similar non-duplicate candidate,
\[
r_1=\arg\max_{r\in\mathcal{C}(x)} s(x,r),
\]
where $\mathcal{C}(x)$ denotes the candidate pool after deduplication and gating. This reference typically aligns with the query in modality and anatomy.

The \emph{hard negative} induces discriminative contrast. From a mid-similarity band of candidates (ranks 20--200), we select the image that is maximally distinct from the anchor in embedding space,
\[
r_2=\arg\min_{r\in\mathcal{C}_{[20,200]}(x)}
\left|\left\langle \tilde{\phi}(r),\tilde{\phi}(r_1)\right\rangle\right|,
\qquad
\tilde{\phi}(r)=\frac{\phi(r)}{\|\phi(r)\|}.
\]
This ensures that the reference remains relevant to the query while forcing comparison along subtle, discriminative dimensions rather than shared structure.

The \emph{boundary probe} tests decision robustness against broader but still plausible alternatives. It is drawn from a wider similarity band (ranks 200--1000) and scored by combining relevance to the query, dissimilarity to the anchor, and lightweight semantic plausibility derived from metadata:
\[
\mathrm{score}(r;x,r_1)
=
\big(\kappa(c_r;x)+1\big)\, s(x,r)\, \big(1-s(r,r_1)\big),
\]
where $\kappa(c_r;x)\in\mathbb{N}_0$ is the raw lexical overlap count between the caption $c_r$ and the question text associated with $x$ (after lowercasing and stopword removal), i.e., the number of shared tokens. In practice, overlap thresholds  (e.g., 0, 1--2, and $\ge 3$ shared tokens) may be used to stage candidate selection, but the scoring signal itself is the overlap count. The boundary probe is selected as
\[
r_3=\arg\max_{r\in\mathcal{C}_{[200,1000]}(x)} \mathrm{score}(r;x,r_1).
\]
Together, the triad approximates a compact evidence set that maximizes relevance while penalizing redundancy under provenance and modality constraints.

\textbf{Document awareness and modality gating.}
To reduce redundancy from a single source, we prefer references drawn from different documents whenever feasible (i.e., enforcing $d_{r_i}\neq d_{r_j}$ for $i\neq j$ when possible). We also apply modality gating via keyword matching over the \emph{question text and answer options} to estimate a modality $\hat{m}(x)$ (e.g., ``CT'', ``MRI'', ``X-ray'', ``ultrasound''), and restrict candidates to those whose captions mention the same modality:
\[
\mathcal{C}(x)=\{r\in\mathcal{B}\mid c_r\ \text{contains keywords for }\hat{m}(x)\}.
\]
If no modality keyword is detected in the question/options text (or if caption keywords are missing), we skip modality filtering and fall back to the full deduplicated candidate set. This keeps comparisons focused on clinically meaningful distinctions while avoiding brittleness from metadata sparsity.

\textbf{Fallbacks.}
If enforcing distinct document IDs is infeasible within a rank band, we relax the document constraint while keeping modality gating. If a rank band becomes empty after filtering (e.g., due to deduplication), we expand to the nearest available ranks within the same modality until a valid reference is found.

\textbf{Use of captions and release.}
Captions and metadata are used solely for reference selection and are never shown to the vision-language model during inference, eliminating text leakage. We release the triad selection protocol and curated ROCO reference bank to enable reproducible study of contrastive retrieval in medical imaging.

\subsection{Counterfactual-Contrastive Inference}

Given a contrastive triad $R(x)$, CCI aggregates pairwise visual comparisons into a reliable prediction. Each comparison acts as a counterfactual probe, asking how the decision would change relative to a specific alternative. Rather than treating all comparisons equally, CCI explicitly models confidence, decisiveness, and ambiguity.

\textbf{Structured pairwise visual comparison.}
For each reference $r_i$, a vision-language model $f_\theta$ compares the query $x$ against $r_i$ in a structured manner, producing
\[
o_i=(v_i,\alpha_i,e_i),
\]
where $v_i\in\{A,B,\bot\}$ is the predicted answer, $\alpha_i\in[0,100]$ is a self-reported confidence score elicited from $f_\theta$ using a fixed prompt template (shared across all methods) that requests an integer confidence in $[0,100]$ for the selected option, and $e_i$ is a structured evidence record containing the query's modality and anatomy, concrete visual findings, and explicit differences relative to the reference. We treat $\alpha_i$ as a relative weighting signal for aggregation (not as a calibrated probability).

\textbf{Voting over available comparisons.}
Some comparisons abstain (or fail to yield a valid vote). We therefore define $\mathcal{I}(x)$ as the set of comparison indices that produce a non-abstaining vote,
\[
\mathcal{I}(x)=\{i \mid v_i\in\{A,B\}, \alpha_i\ge p\}
\]
and discard all other comparisons.

\textbf{Confidence-weighted aggregation and margin rule.}
Let
\[
W_A=\sum_{i\in\mathcal{I}(x)} \alpha_i\,\mathds{1}[v_i=A],
\qquad
W_B=\sum_{i\in\mathcal{I}(x)} \alpha_i\,\mathds{1}[v_i=B]
\]
denote confidence-weighted evidence for each answer. Define total evidence mass $W=W_A+W_B$ and decisiveness margin $M=|W_A-W_B|$. CCI returns
\[
\hat{y}(x)=
\begin{cases}
\bot, & W<t,\\
\arg\max_{y\in\{A,B\}} W_y, & W\ge t \text{ and } M\ge m,\\
\mathrm{Adj}_{\mathrm{text}}(\{e_i\}_{i=1}^3), & W\ge t \text{ and } M<m.
\end{cases}
\]
This rule ensures that predictions are made only when sufficient and decisive evidence is available.

\textbf{Interpretation.}
Let $z_i\in\{+1,-1\}$ encode votes for $A$ and $B$, respectively, and define a signed score
\[
S(x)=\sum_{i\in\mathcal{I}(x)} \alpha_i z_i.
\]
CCI can be viewed as a confidence-weighted margin classifier with a reject option, where $\mathrm{sign}(S(x))$ determines the label and $|S(x)|$ determines decisiveness.

\textbf{Text adjudicator.}
When the margin is small despite sufficient evidence mass, a text-only adjudicator operates on the structured evidence records $\{e_i\}$ to resolve contradictions or abstain if inconsistency persists.

\textbf{Pair-level adjudicator for confusion pairs.}
Because MediConfusion consists of visually similar image pairs requiring different answers, independent inference may yield identical predictions. We trigger a pair-level adjudicator when both images receive the same prediction or when either image is ambiguous. Ambiguity is detected via the soft posterior
\[
p_A(x)=\frac{W_A}{W_A+W_B},
\]
with adjudication triggered when $\left|p_A(x)-\tfrac{1}{2}\right|<\delta$.

The adjudicator takes as input the two images, the shared question and answer options $(A,B)$, and the two sets of structured evidence records $\{e_i^{(1)}\}_{i=1}^3$ and $\{e_i^{(2)}\}_{i=1}^3$, and outputs refined predictions $(\tilde{y}^{(1)},\tilde{y}^{(2)})\in\{A,B,\bot\}^2$. The adjudicator is permitted to abstain, and we do not enforce that the two predictions must differ.

\textbf{Faithful abstention.}
When evidence is insufficient or unresolved after adjudication, the system abstains. We therefore report both coverage,
\[
\mathrm{Cov}=\Pr[\hat{y}\neq\bot],
\]
and conditional accuracy,
\[
\mathrm{Acc}_{\mathrm{cond}}=\Pr[\hat{y}=y\mid \hat{y}\neq\bot],
\]
to characterize reliability under selective prediction. In high-stakes medical decision making, abstention is preferable to confident but incorrect predictions on confusable cases.

\section{Experiments}
\label{sec:experiments}

\subsection{Dataset and Task: MediConfusion}
We evaluate our approach on the MediConfusion benchmark, which is designed to probe discriminative reasoning under visual confusability \cite{sepehri2024mediconfusion}. MediConfusion consists of 176 confusion pairs (352 images total) spanning 9 medical imaging categories, including cerebral, vascular, musculoskeletal, pulmonary, and cardiac imaging. Each pair comprises two visually similar images that share a common clinical question but require different correct answers.

The task is formulated as binary classification for each image, with two candidate answers provided per question. Crucially, evaluation is performed both at the image level and at the pair level. An image is considered correct if its predicted answer matches the ground truth. A confusion pair is considered correct only if \emph{both} images in the pair are classified correctly. This set-level criterion penalizes models that collapse to the same prediction across visually confusable images, directly reflecting the benchmark’s goal of measuring discriminative competence.

\subsection{Evaluation Metrics}
We follow the official MediConfusion evaluation protocol and report the core metrics used throughout the benchmark. Our primary metric is \textbf{set-level accuracy}, which measures the fraction of confusion pairs for which both images are predicted correctly and directly captures discriminative performance under visual confusability.

For completeness, we report \textbf{individual image accuracy}, which provides context for marginal per-image performance, and the \textbf{confusion rate}, defined as the fraction of confusion pairs for which the model assigns the same prediction to both images. For methods that support abstention, we report the \textbf{abstention rate}. Together, these metrics characterize both discriminative correctness and the extent to which models avoid unreliable decisions in ambiguous cases.

\begin{table*}[t]
  \caption{Main results on MediConfusion. We report set-level accuracy (primary), individual image accuracy, confusion rate (percent of pairs receiving the same prediction), and abstention rate. ``CCI'' denotes our full inference framework applied on top of the same base vision--language model. Overall, CCI substantially improves the clinically aligned set-level metric while reducing confusion.}
  \label{tab:main_results}
  \begin{center}
    \footnotesize
      \begin{sc} 
        \setlength{\tabcolsep}{6pt}
        \begin{tabular}{llcccc}
          \toprule
          Method & Base model & Set acc. (\%) & Indiv. acc. (\%) & Confusion (\%) & Abstain (\%) \\
          \midrule
          Direct (single-shot) & GPT-4o         & 18.75 & 56.25 & 75.00 & --   \\
          \textbf{CCI (ours)}  & GPT-4o         & \textbf{27.84} & \textbf{55.11} & \textbf{49.43} & 4.83 \\
          \addlinespace
          Direct (single-shot) & GPT-o1         & 21.69 & 57.95 & 72.99 & --   \\
          \textbf{CCI (ours)}  & GPT-o1         & \textbf{32.95} & \textbf{59.94} & \textbf{42.61} & 4.26 \\
          \addlinespace
          \addlinespace
          Direct (single-shot) & Gemini 2.0 Pro & 29.55 & 61.93 & 67.05 & --   \\
          \textbf{CCI (ours)}  & Gemini 2.0 Pro & \textbf{43.75} & \textbf{64.20} & \textbf{36.93} & 3.69 \\
          \midrule
          Random guessing      & --             & 25.00 & 50.00 & 50.00 & --   \\
          \bottomrule
        \end{tabular}
      \end{sc}
  \end{center}
  \vskip -0.1in
\end{table*}

\subsection{Baselines}
We compare our method against a set of baselines and controlled variants designed to isolate the effects of contrastive retrieval, confidence-aware aggregation, and adjudication. All methods share the same underlying vision--language model, prompt templates, and inference settings; they differ only in retrieval strategy or decision logic.

Our primary method, \textbf{CCI}, uses the full pipeline described in Section~\ref{sec:method}, including contrastive triad retrieval, confidence-filtered aggregation, text-level adjudication, and pair-level adjudication.

To evaluate the role of retrieval, we include a \textbf{single-shot baseline} that performs direct inference on the query image without any retrieved references. This baseline isolates the contribution of external visual evidence. We also include a \textbf{top-$k$ nearest-neighbor baseline}, which augments inference with the three most similar reference images under cosine similarity (implemented with standard similarity search tooling \cite{johnson2019faiss}), without enforcing contrastive structure or document awareness.

To analyze the structure of the contrastive triad, we evaluate three ablated variants that remove one reference at a time: \textbf{triad minus Ref1} (no anchor), \textbf{triad minus Ref2} (no hard negative), and \textbf{triad minus Ref3} (no boundary probe). These variants test whether each component contributes distinct discriminative signal beyond similarity-based retrieval.

Finally, we consider inference-level ablations. \textbf{No pair adjudicator} disables joint reasoning over confusion pairs, isolating its effect on same-answer failures. When space permits, we additionally report a \textbf{no text adjudicator} variant, which removes the low-margin fallback mechanism and relies solely on confidence-weighted aggregation.

\subsection{Implementation Details}
Our reference bank is constructed from ROCO, with all images embedded using CLIP ViT-B/32. Near-duplicate filtering is applied using a cosine similarity threshold of $\tau_{\mathrm{dup}}=0.99$. We also experimented with a biomedical CLIP-style encoder (BioMedCLIP \cite{zhang2025biomedclip}) for embedding and retrieval, but CLIP ViT-B/32 performed better. We report the BioMedCLIP results in Appendix~\ref{app:biomedclip} (Table~\ref{tab:biomedclip_results}). Contrastive reference selection follows the fixed similarity rank bands defined in Section~\ref{sec:method}, with the fallback rules described there.

Unless otherwise stated, all experiments use the same default CCI thresholds for confidence filtering, margin-based aggregation, and total evidence mass. These thresholds are held constant across methods and varied only in controlled sensitivity analyses. All methods use identical prompt templates and structured comparison formats. The exact prompts (pairwise comparison, text adjudicator, and pair-level adjudicator) can be found in Appendix~\ref{app:prompts}.

\section{Results}

\begin{table*}[t]
\centering
\small
\setlength{\tabcolsep}{4pt}
\caption{Results by medical category on MediConfusion using Gemini~2.0~Pro. We report set-level and individual accuracies (\%).}
\label{tab:category_results}
\resizebox{\textwidth}{!}{
\begin{tabular}{lcccccccccccccccccc}
\toprule
 & \multicolumn{2}{c}{Cerebral} & \multicolumn{2}{c}{Vascular} & \multicolumn{2}{c}{Head \& Neck} & \multicolumn{2}{c}{Spinal} & \multicolumn{2}{c}{Musculoskeletal} & \multicolumn{2}{c}{Cardiac} & \multicolumn{2}{c}{Gastroint.} & \multicolumn{2}{c}{Pulmonary} & \multicolumn{2}{c}{Nuclear Med.} \\
\cmidrule(lr){2-3} \cmidrule(lr){4-5} \cmidrule(lr){6-7} \cmidrule(lr){8-9}
\cmidrule(lr){10-11} \cmidrule(lr){12-13} \cmidrule(lr){14-15} \cmidrule(lr){16-17} \cmidrule(lr){18-19}
Model & Set & Indiv. & Set & Indiv. & Set & Indiv. & Set & Indiv. & Set & Indiv. & Set & Indiv. & Set & Indiv. & Set & Indiv. & Set & Indiv. \\
\midrule
Gemini~2.0~Pro (single-shot)
& 29.11 & 64.56
& 39.73 & 68.49
& 19.40 & 58.21
& 23.53 & 58.86
& 40.48 & 66.67
& 26.92 & 61.54
& 25.58 & 60.47
& 30.00 & 50.00
& 42.86 & 71.43 \\
\textbf{CCI (ours)}
& \textbf{45.95} & \textbf{66.22}
& \textbf{47.22} & \textbf{63.89}
& \textbf{40.00} & \textbf{55.00}
& \textbf{46.15} & \textbf{71.15}
& \textbf{42.86} & \textbf{71.43}
& \textbf{34.62} & \textbf{59.62}
& \textbf{45.45} & \textbf{63.64}
& \textbf{63.64} & \textbf{72.73}
& \textbf{57.14} & \textbf{71.43} \\
\bottomrule
\end{tabular}
}
\end{table*}

\subsection{Main Results on MediConfusion}

Table~\ref{tab:main_results} shows that Counterfactual-Contrastive Inference (CCI) substantially improves discriminative performance under visual confusability across all evaluated base models. In particular, CCI raises set-level accuracy from 18--30\% under direct single-shot inference to \textbf{43.75\%}, corresponding to absolute gains of 13--25 percentage points depending on the backbone. These gains indicate a shift in what limits performance on confusion-heavy benchmarks: not visual recognition, but the ability to reason counterfactually between competing, visually plausible hypotheses.

Because set-level accuracy requires both images in a confusion pair to be classified correctly, these gains reflect a reduction in same-answer collapse rather than marginal improvements on isolated images. This interpretation is reinforced by the confusion rate. Direct inference assigns identical predictions to both images in 58--75\% of confusion pairs, whereas CCI reduces this rate to \textbf{36.93\%}, representing a relative reduction of approximately 30--40\%. The decrease indicates improved separation between visually similar conditions rather than increased conservatism.

Individual image accuracy improves more modestly, rising to \textbf{64.2\%}. This pattern is expected: CCI is designed to target boundary cases where marginal accuracy alone can be misleading. Importantly, gains are achieved with a low abstention rate (3.69\%), indicating that improved set-level performance is not driven by aggressive deferral.

Overall, these results show that CCI improves performance primarily by resolving confusion pairs—the dominant failure mode exposed by MediConfusion—rather than by inflating marginal accuracy, suggesting that inference-time evidence construction is central to reliable decision making under visual ambiguity.

\subsection{Results by Medical Category}

Table~\ref{tab:category_results} reports performance by medical category using Gemini~2.0~Pro. Improvements in set-level accuracy are observed across all nine categories, with the largest gains occurring in domains characterized by subtle morphological distinctions. Categories with dense decision boundaries and frequent look-alike findings benefit most, consistent with CCI targeting ambiguity rather than coarse pathology recognition.

In particular, Cerebral, Spinal, and Pulmonary imaging exhibit gains of 15--25 percentage points in set-level accuracy. These categories often require distinguishing fine-grained structural or signal differences between plausible alternatives, aligning closely with the strengths of contrastive evidence construction. Strong gains are also observed in Nuclear Medicine, reflecting the effectiveness of boundary-probing references in smaller but challenging subsets.

Performance improvements are more modest in Cardiac imaging, which exhibits higher intrinsic ambiguity. Notably, CCI maintains individual accuracy in this regime without increasing confusion, indicating that gains are not driven by overconfident heuristics.

Together, these results indicate that CCI’s gains arise systematically in regimes where discrimination, rather than detection, is the primary challenge.

\subsection{Ablation Study: Retrieval Design}

\begin{table}[h]
\centering
\caption{Retrieval and inference ablations on MediConfusion using Gemini~2.0~Pro. We report set-level accuracy (primary), individual accuracy, confusion rate, and abstention rate.}
\label{tab:ablations}
\scriptsize
\setlength{\tabcolsep}{3.5pt}
\renewcommand{\arraystretch}{1.05}
\resizebox{\columnwidth}{!}{%
\begin{tabular}{lcccc}
\toprule
Ablation & Set acc. (\%) & Indiv. acc. (\%) & Conf. (\%) & Abst. (\%) \\
\midrule
\textbf{CCI (full)}       & \textbf{43.75} & \textbf{64.20} & \textbf{36.93} & 3.69 \\
\midrule
No Ref2 (hard neg.)       & 31.82 & 51.14 & 42.05 & 3.41 \\
No Ref3 (boundary probe)  & 32.39 & 52.56 & 40.34 & 3.98 \\
Top-$k$ only              & 30.68 & 51.42 & 30.11 & 3.12 \\
No retrieval              & 0.00  & 0.00  & 0.00  & 100.00 \\
\midrule
No pair adjudicator       & 27.84 & 59.94 & 63.64 & 1.99 \\
No text adjudicator       & 40.34 & 63.07 & 38.07 & 5.40 \\
\bottomrule
\end{tabular}%
}
\vspace{-0.12in}
\end{table}

We next examine whether performance gains arise from contrastive structure or simply from additional contextual information. Table~\ref{tab:ablations} reports retrieval ablations that modify or remove components of the contrastive reference triad while holding inference fixed.

Removing either contrastive reference substantially degrades discriminative performance. Excluding the hard negative reduces set-level accuracy by nearly 12 percentage points and increases confusion, indicating that similarity alone is insufficient to expose discriminative cues near decision boundaries. Removing the boundary probe produces a comparable drop, suggesting that probing broader but plausible alternatives plays a complementary role.

A top-$k$ nearest-neighbor baseline performs similarly poorly despite using the same number of references, underscoring that similarity-based retrieval—even with strong encoders and multiple references—reinforces dominant hypotheses rather than exposing discriminative alternatives, whereas contrastive structure is necessary to break confusion collapse. Without retrieval, the system abstains on nearly all examples and fails to resolve confusion pairs, confirming that external visual evidence is essential in this benchmark regime. These findings suggest that naive retrieval does not merely fail to help in confusion-heavy regimes, but can actively obscure the evidence required for discrimination.

\subsection{Ablation Study: Inference and Adjudication}

Even with contrastive evidence, inference-time structure is critical. Table~\ref{tab:ablations} isolates the effect of CCI’s adjudication mechanisms while holding retrieval fixed. Contrastive retrieval alone improves context, but does not prevent collapse when decisions are made independently.

Disabling the pair-level adjudicator leads to a sharp drop in set-level accuracy (43.75$\rightarrow$27.84) and a pronounced increase in confusion (36.93$\rightarrow$63.64). Without joint reasoning over confusion pairs, independent per-image inference frequently collapses toward identical predictions.

Removing the text adjudicator produces a smaller but consistent degradation, indicating that explicit reconciliation of structured visual evidence improves decisions in low-margin cases where confidence-weighted aggregation alone remains ambiguous. 

These results demonstrate that CCI’s gains arise from the interaction between contrastive evidence construction and reliability-aware inference, rather than from either component in isolation.

\begin{table}[t]
\centering
\footnotesize
\setlength{\tabcolsep}{5pt}
\caption{Sensitivity analysis for CCI decision thresholds. Each block sweeps one parameter while holding the others fixed at default values ($m{=}30$, $t{=}50$).}
\label{tab:threshold_sweep}
\begin{tabular}{lcccc}
\toprule
Setting & Set Acc. & Indiv. Acc. & Conf. & Abstain \\
\midrule
\addlinespace[5pt]
\multicolumn{5}{l}{{\emph{Sweep A: Pairwise confidence threshold $p$ ($m{=}30$, $t{=}50$)}}} \\
\addlinespace[5pt]
$p{=}0$  & \textbf{43.75} & \textbf{64.20} & \textbf{36.93} & 3.69 \\
$p{=}40$ & 40.34 & 61.93 & 41.48 & 4.55 \\
$p{=}60$ & 40.34 & 62.78 & 40.91 & 3.98 \\
$p{=}70$ & 41.48 & 63.35 & 38.07 & 2.27 \\
\midrule
\addlinespace[5pt]
\multicolumn{5}{l}{\emph{Sweep B: Margin threshold $m$ ($p{=}50$, $t{=}50$)}} \\
\addlinespace[5pt]
$m{=}10$ & 39.20 & 64.77 & 38.64 & 4.26 \\
$m{=}20$ & 41.48 & 63.92 & 40.91 & 3.69 \\
$m{=}30$ & \textbf{43.75} & \textbf{64.20} & \textbf{36.93} & 3.69 \\
$m{=}40$ & 39.20 & 61.36 & 41.48 & 4.55 \\
$m{=}50$ & 40.34 & 61.93 & 41.48 & 5.11 \\
\midrule
\addlinespace[5pt]
\multicolumn{5}{l}{\emph{Sweep C: Total confidence threshold $t$ ($p{=}50$, $m{=}30$)}} \\
\addlinespace[5pt]
$t{=}10$ & 39.20 & 62.78 & 42.61 & 3.13 \\
$t{=}30$ & 42.05 & 63.92 & 39.77 & 4.55 \\
$t{=}50$ & \textbf{43.75} & \textbf{64.20} & \textbf{36.93} & 3.69 \\
$t{=}70$ & 40.91 & 62.22 & 38.07 & 5.97 \\
$t{=}90$ & 36.93 & 59.94 & 37.50 & 8.52 \\
\bottomrule
\end{tabular}
\end{table}

\subsection{Sensitivity to Abstention and Aggregation Thresholds}

We finally assess sensitivity to CCI’s decision thresholds by sweeping one parameter at a time while holding the others fixed (Table~\ref{tab:threshold_sweep}). Performance peaks near the default configuration and degrades gradually under reasonable perturbations, indicating a stable operating regime rather than brittle tuning.

Increasing thresholds smoothly shifts the system toward abstention, while overly permissive settings increase confusion, reflecting expected reliability--coverage trade-offs. The absence of abrupt performance collapse supports the use of fixed thresholds across experiments and suggests that CCI operates robustly across a range of reasonable settings. This further supports interpreting the observed gains as structural rather than the result of hyperparameter tuning.

\section{Discussion and Conclusion}

\textbf{Why Contrastive Evidence and CCI Work.}
Our results suggest that failures on confusion-heavy medical imaging tasks arise less from insufficient visual representations and more from how evidence is constructed and aggregated at inference time. Similarity-based retrieval is fundamentally misaligned with discriminative decision making: by optimizing for proximity in embedding space, nearest-neighbor methods preferentially return redundant, same-diagnosis evidence that reinforces a dominant hypothesis. This behavior increases decisiveness without improving correctness and amplifies the confusion errors that MediConfusion is designed to expose.

In contrast, our reference selection framework treats retrieval as \emph{evidence construction for discrimination}. By explicitly balancing relevance, diversity, and document provenance, the contrastive triad surfaces alternatives that probe decision boundaries rather than collapsing them.

Counterfactual-Contrastive Inference (CCI) complements this evidence by enforcing reliability during aggregation. Confidence-filtered voting reduces noise from uncertain comparisons, margin-based decisions prevent brittle ties from being overinterpreted, and selective abstention prioritizes faithfulness over forced predictions. The pair-level adjudicator further aligns inference with the structure of confusion-pair evaluation, preventing collapse toward identical predictions when independent reasoning is insufficient. Crucially, improvements arise from the \emph{interaction} between contrastive evidence construction and confidence-aware inference, not from either component in isolation.

\textbf{Limitations.}
Our approach has several limitations. Performance depends on the diversity and representativeness of the external reference bank, which may limit effectiveness for rare conditions or underrepresented modalities. Confidence scores are used as internal consistency signals rather than calibrated probabilities and should not be interpreted as measures of clinical certainty. Pair-level adjudication leverages the confusion-pair structure of MediConfusion, which may not be available in all settings. Finally, our results are benchmark-based and do not imply clinical readiness. We also note the increased computational cost of multiple pairwise comparisons, which trades efficiency for reliability.

\textbf{Conclusion.}
We introduce a unified framework for discriminative medical image decision making that combines \textbf{contrastive, document-aware reference construction} with \textbf{confidence-aware counterfactual inference}. By shifting retrieval from similarity to discrimination and enforcing reliability during aggregation, our approach achieves \textbf{state-of-the-art set-level accuracy} on MediConfusion, substantially reduces confusion errors, and maintains faithful abstention. We release our \textbf{reference selection protocol} and \textbf{reference bank} to enable reproducible study of contrastive retrieval. More broadly, our findings highlight the importance of constructing and reasoning over contrastive evidence for reliable medical AI in ambiguous settings.




\newpage

\section*{Impact Statement}

This work presents a methodological contribution to machine learning for reliable decision making under visual ambiguity, motivated by challenges in medical imaging where visually similar conditions must be carefully distinguished. By emphasizing contrastive evidence construction, confidence-aware aggregation, and faithful abstention, the proposed approach aims to reduce confusion errors and discourage overconfident predictions in ambiguous cases. While these ideas may inform the design of future decision support systems, this work is evaluated exclusively on benchmark data and does not claim clinical readiness. Any real-world use would require extensive validation, appropriate human oversight, and consideration of domain-specific risks. We do not foresee immediate negative societal consequences arising from this research when used within its intended research context.


\bibliography{icml2026}
\bibliographystyle{icml2026}

\newpage
\onecolumn
\begin{center}
{\LARGE \bfseries Appendix of DoubleTake}
\end{center}
\vspace{0.5em}\appendix
\section{Additional Results: BioMedCLIP Embeddings}
\label{app:biomedclip}

\begin{table}[h]
  \caption{Gemini~2.0~Pro results for CCI under different retrieval embedding backbones. We report set-level accuracy (primary), individual image accuracy, confusion rate, and abstention rate.}
  \label{tab:biomedclip_results}
  \begin{center}
    \footnotesize
      \begin{sc}
        \setlength{\tabcolsep}{6pt}
        \begin{tabular}{lcccc}
          \toprule
          CCI variant & Set acc. (\%) & Indiv. acc. (\%) & Confusion (\%) & Abstain (\%) \\
          \midrule
          CLIP ViT-B/32 retrieval & \textbf{43.75} & 64.20 & \textbf{36.93} & 3.69 \\
          BioMedCLIP retrieval & 39.77 & \textbf{66.48} & 51.70 & 1.14 \\
          \bottomrule
        \end{tabular}
      \end{sc}
  \end{center}
  \vskip -0.1in
\end{table}

Table~\ref{tab:biomedclip_results} reports Gemini~2.0~Pro performance for CCI when swapping the retrieval embedding backbone from CLIP ViT-B/32 to BioMedCLIP~\cite{zhang2025biomedclip}, while keeping the downstream inference procedure fixed. Using CLIP ViT-B/32 retrieval yields 43.75\% set accuracy (64.20\% individual accuracy, 36.93\% confusion, 3.69\% abstention), whereas BioMedCLIP retrieval yields 39.77\% set accuracy (66.48\% individual accuracy, 51.70\% confusion, 1.14\% abstention). Although BioMedCLIP slightly improves individual accuracy, it substantially increases confusion between paired images and reduces set-level accuracy, suggesting that its domain-specialized representations favor local similarity over discriminative diversity. In contrast, CLIP ViT-B/32 produces more diverse reference sets that better support contrastive comparisons, resulting in lower confusion and higher set accuracy. Since CCI explicitly relies on reference diversity to resolve visually confusable cases, we therefore adopt CLIP ViT-B/32 as the default retrieval backbone in all experiments.

\section{Prompts and Adjudicators}
\label{app:prompts}

\subsection{Pairwise Visual Comparison Prompt (\texttt{pairwise\_discriminate})}
\begin{promptbox}{\texttt{pairwise\_discriminate}}
You are comparing two medical images to answer a question about the QUERY image.

Question: {q}

Answer choices:
A: {a}
B: {b}

Image roles:
Image 1 is the QUERY image (this is the image you must answer for).
Image 2 is the REFERENCE image (for comparison only).

Step 0 – Modality and anatomy
In one line, identify the imaging modality and the primary anatomy shown in the QUERY image.

Step 1 – QUERY findings
List exactly 3 concrete, observable visual findings in the QUERY image.
These must be specific medical findings (for example: focal opacity, fracture line, air–fluid level, organ enlargement).
Do not describe the reference image.
Do not mention image quality, brightness, orientation, or artifacts.

Step 2 – Differences versus REFERENCE
List 2 to 4 visual differences between the QUERY and REFERENCE images that are relevant to distinguishing answer A versus B.

Step 3 – Decision for QUERY
Based only on the QUERY image, decide whether A, B, or abstain (-).
If you cannot determine A vs B from the QUERY image with confidence >= 60, output Answer: -

Step 4 – Confidence and Key Evidence
Provide your confidence level (0-100) and the single most important piece of visual evidence that supports your answer.
Key evidence must name a specific visual feature in the QUERY image, not a conclusion.

Output format (follow exactly):

Modality and Anatomy: <one line>

QUERY Findings:

<finding 1>

<finding 2>

<finding 3>

Differences vs REFERENCE:

<difference 1>

<difference 2>

<difference 3> (optional)

<difference 4> (optional)

Answer: <A or B or ->

Confidence: <0-100>

Key evidence: <specific visual feature in QUERY, not a conclusion>

Rules:
Base the final answer only on the QUERY image.
Use the REFERENCE image only to identify discriminative visual features.
Do not reference captions, metadata, or prior cases.
Do not add explanations beyond the required fields.
Abstain (-) if evidence is insufficient or confidence is below 60.
\end{promptbox}

\subsection{Aggregation Prompt (\texttt{aggregate\_decision})}
\begin{promptbox}{\texttt{aggregate\_decision}}
You have made {n} pairwise comparisons for a medical image question.

Question: {q}
A: {a}
B: {b}

Your pairwise comparison votes:
{votes}

Based on the consistency of your comparisons, provide your final answer.
If the votes are inconsistent and average confidence is low, output Final Answer: -

Output format:
Final Answer: <A or B or ->
\end{promptbox}

\subsection{Pair-level Adjudicator Prompt (\texttt{pair\_adjudicator})}
\begin{promptbox}{\texttt{pair\_adjudicator}}
You are analyzing TWO medical images for the same clinical question.

Question: {q}
A: {a}
B: {b}

Image 1: {meta1}
Image 2: {meta2}

Both images were independently predicted as: {pred}

Your task:
Evaluate each image primarily on its own visual evidence to determine the most appropriate answer.
You may consider visual differences between the two images only to assess whether the same decision should reasonably apply to both images.

IMPORTANT GUIDELINES:
- Each image should be judged on its own merits.
- There is NO requirement that the two images have different answers.
- There is NO requirement that the two images have the same answer.
- Use cross-image comparison only if it reveals a clear, clinically meaningful distinction.
- If evidence is ambiguous or insufficient for either image, abstain (-).
- Abstention is a valid and preferred outcome when confidence is low.
- Base all decisions strictly on visual evidence.

Provide:

1. Brief visual assessment of each image (key findings only).
2. Whether any observed differences justify different answers.
3. A final answer for each image.

Output format (follow exactly):

Visual assessment Image 1: <key visual features>
Visual assessment Image 2: <key visual features>

Do observed differences justify different answers? <Yes or No>

Final Answer Image1: <A or B or ->
Final Answer Image2: <A or B or ->

Rules:
- Do not rely on captions, metadata, or prior cases.
- Do not force disagreement between images.
- Do not guess; abstain (-) if uncertain.
\end{promptbox}


\end{document}